\documentclass[a4paper, 10pt, conference]{ieeeconf}
\IEEEoverridecommandlockouts 

\usepackage{lipsum}
\usepackage{amsmath}
\usepackage{graphicx}
\usepackage{multirow}
\usepackage{subcaption}
\title{\LARGE \bf
MLOD: A multi-view 3D object detection based on robust feature fusion method
}
\author{Jian Deng$^{1}$ and Krzysztof Czarnecki$^{2}$
\thanks{$^{1}$Jian Deng is with David R. Cheriton School of Computer Science, University of Waterloo,  200   University   Avenue   West,   Waterloo,   Canada 
        {\tt\small jian.deng@uwaterloo.ca}}%
\thanks{$^{2}$Krzysztof  Czarnecki is  with  the  Department  of  Electrical  and  Computer  Engineering,  University  of  Waterloo,  200  University Avenue  West,  Waterloo,  Canada
        {\tt\small Canadakczarnec@gsd.uwaterloo.ca}}%
}

\begin{document}

\maketitle
\thispagestyle{empty}
\pagestyle{empty}

\begin{abstract}

This paper presents Multi-view Labelling Object Detector (MLOD). The detector takes an RGB image and a LIDAR point cloud as input and follows the two-stage object detection framework \cite{girshick2015fast} \cite{ren2015faster}. A Region Proposal Network (RPN) generates 3D proposals in a Bird's Eye View (BEV) projection of the point cloud. The second stage projects the 3D proposal bounding boxes to the image and BEV feature maps and sends the corresponding map crops to a detection header for classification and bounding-box regression. Unlike other multi-view based methods, the cropped image features are not directly fed to the detection header, but masked by the depth information to filter out parts outside 3D bounding boxes. The fusion of image and BEV features is challenging, as they are derived from different perspectives. We introduce a novel detection header, which provides detection results not just from fusion layer, but also from each sensor channel. Hence the object detector can be trained on data labelled in different views to avoid the degeneration of feature extractors. MLOD achieves state-of-the-art performance on the KITTI 3D object detection benchmark. Most importantly, the evaluation shows that the new header architecture is effective in preventing image feature extractor degeneration.       

\end{abstract}


\section{Introduction}
3D object detection is crucial for a safe and robust mobile robotic system, like an autonomous vehicle. It allows such a system to track and predict the motion of objects by providing classification and localization of physical objects. LIDAR is a common sensor in autonomous driving, used to measure 3D structure of the surrounding environment. Due to the unique characteristics of point cloud data, 2D object detection methods have not transferred well to the detection of 3D objects using LIDAR.

In this work, 3D point cloud data is represented in the form of a birds-eye view (BEV) map, which contains multiple channels of height and density information. Several multi-view 3D object detectors with BEV map as input exist \cite{ku2018joint} \cite{chen2017multi}. These methods apply convolutional neural networks (CNN) to the BEV map and RGB image data, and use the resulting fused features to detect objects. In these methods, multi-view detection networks are trained in an end-to-end fashion. During training, object proposals are labeled according to Intersection-over-Union (IoU) in BEV. However the IoU of proposals is different in top-down view and front view, and as a result the labelled data becomes `noisy' for the image channel. Consequently, the negative samples with high IoU in front view lead to the deterioration of image feature extractor. 

\begin{figure}[ht]
\centering
\includegraphics[width=200pt, height = 150pt]{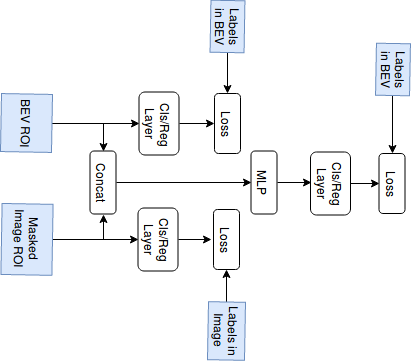}
\caption{The multi-view header architecture diagram}
\label{fig:multi-view_header} 
\end{figure}

We propose a Multi-view Labelling Object Detector (MLOD) to address this problem. The main contribution of this paper is listed as follows:
\begin{itemize}
    \item We propose a foreground mask layer, which exploits the projected depth map in front view to select the foreground image features within a 3D bounding box proposal.
    \item We propose a multi-view detection header (Fig. \ref{fig:multi-view_header}) which has output not only from fusion layer, but also from each sensor channel. This design enables our detection network to be trained on the samples labelled based on IoU in the view of each channel. 
\end{itemize}

Our paper is organized as follows. We first present an overview of the current 3D object detectors that use both LIDAR point cloud and image data in Section \ref{sec2}. Section \ref{sec3} outlines the proposed detection network architecture. Section \ref{sec4} gives the implementation details, followed by the experimental results in Section \ref{sec5}. Finally,  we conclude in Section \ref{sec6}. 

\section{Related Works}
\label{sec2}

There are roughly three ways to take advantage of camera and LIDAR for 3D objection detection for autonomous driving: image region proposal, projection-based and multi-view methods.

\begin{figure*}[ht]
\centering
\includegraphics[width=400pt, height = 180pt]{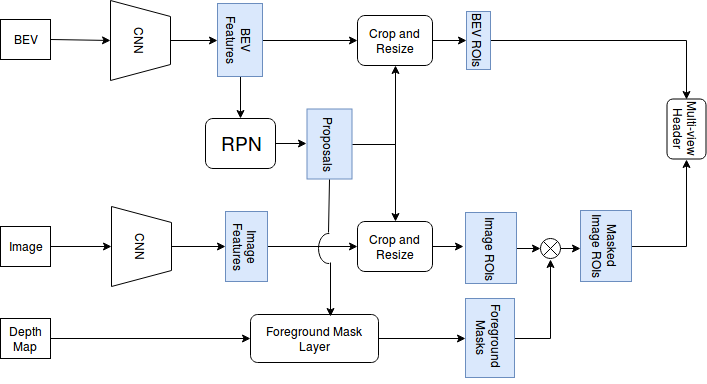}
\caption{Architectural diagram of the proposed method}
\label{fig:main_diagram} 
\end{figure*}

\subsection{Image region proposal methods}
F-PointNet \cite{qi2018frustum} uses an image detection module to provide 2D bounding boxes as proposals. Then the LIDAR points inside the proposals are cropped and fed into an instance segmentation module to select the positive points. Finally, two PointNets \cite{qi2017pointnet} predict the bounding box within the selected LIDAR points. IPOD \cite{yang2018ipod} implements a 2D semantic segmentation network to filter out background LIDAR points. Then it classifies and refines 3D bounding boxes on the remaining foreground points.

\subsection{Projection-based methods}
Liang et al. \cite{liang2018deep} proposed a method that projects image features into BEV and fuses them with the convolutional layers of a LIDAR based detector using a continuous fusion layer. The layer creates a continuous BEV feature map where each pixel in BEV contains the corresponding image information. For each BEV pixel, the detector first finds the k nearest LIDAR points on the BEV map, then obtains the image feature at the continuous coordinates by bilinear interpolation. The interpolated image features and geometry offsets are feed into a multilayer perceptron (MLP). Then the deep continuous fusion networks fuse multi-sensor information by the summation of BEV features and output from MLP.

\subsection{Multi-view based methods}
MV3D \cite{chen2017multi} and AVOD \cite{ku2018joint} are two-stage detectors. The multi-view based methods merge features from BEV map and RGB image to predict the 3D bounding boxes. MV3D only uses BEV maps in its RPN to generate proposals, and AVOD uses both BEV and image views. When small instances (like pedestrians and cyclists) on BEV maps are down-sampled by pooling layers, object features are compressed into one pixel in the final feature map, which is insufficient for the second stage detection. Hence, AVOD-FPN improved MV3D by using pyramid convolution structure in BEV/image feature extractors. In AVOD, features are merged in the refinement phase.

The existing multi-view based approaches tend to rely more on BEV map input rather than RGB images. Two-stage methods usually select the top k proposals to feed into the detection header networks. Due to the different viewpoints, the IoUs with ground-true box in BEV and image view, respectively, are also different. Hence some proposals are labelled as negative samples in BEV, but should be treated as positive in image view (see Fig. \ref{fig:label}). Since the positive/negative samples in the current multi-view neural network architectures are assigned based on the IoU in BEV, some positive samples in image view are labelled as negative ones, and thus the image feature extractor is trained on the `noisy' labels. This problem weakens the performance of the image channel. Therefore the existing multi-view 3D object detectors tend to fail to leverage the image information, and only concentrate on the BEV map. 

\section{The MLOD Architecture}
\label{sec3}
The proposed two-stage neural network architecture is presented in Fig. \ref{fig:main_diagram}. BEV map and RGB image are fed into two convolution neural networks to obtain features. For computational efficiency, we only use the BEV features in RPN to generate 3D proposals. Based on the depth information of the proposals, image features outside 3D proposals are masked by a foreground mask layer. Then the masked image feature map and the BEV feature map are cropped and passed to multi-view header to provide the final classification, localization, and orientation results.

\subsection{BEV Map Preparation \& Feature Extractor}
Similar to \cite{chen2017multi} \cite{ku2018joint}, the six-channel BEV map input is a 2D grid with 0.1 meter resolution, which includes five height channels and a single  density channel. The point cloud is divided into 5 equal slices between [0, 2.5] meters along the normal of the ground plane, and each slice produces a height channel with each grid cell representing the maximum height of points in that cell.

We adopt the U-Net \cite{ronneberger2015u} structure from \cite{ku2018joint} as BEV feature extractor. The encoder part is a VGG-like CNN \cite{simonyan2014very}, but with half of the channels. It  includes CNN layers only up to conv-4. In the decoder part, feature extractors use the conv-transpose operation to up-sample the feature maps. The up-sampled feature maps are fused with corresponding features from encoder via concatenation. Image feature extractor is a VGG16 CNN before pool-5 layer.

\subsection{Foreground Mask Layer}
To correctly capture the image features of the object inside the proposed 3D bounding box, we introduce a foreground masking layer to filter out the foreground features.

\begin{figure}[ht]
\centering
\begin{subfigure}[b]{0.5\textwidth}
\includegraphics[width=250pt, height = 180pt]{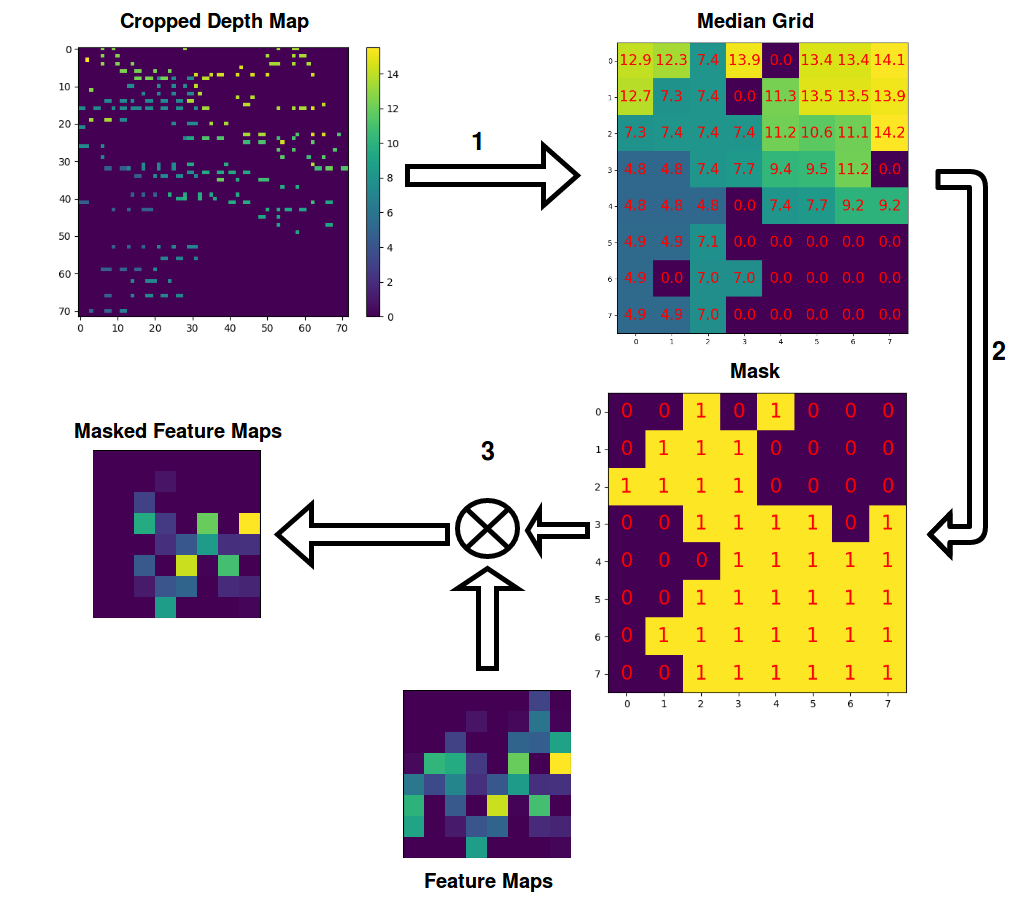}
\caption{}
\label{fig:1}
\end{subfigure}

\begin{subfigure}[b]{0.5\textwidth}
\includegraphics[width=250pt, height = 80pt]{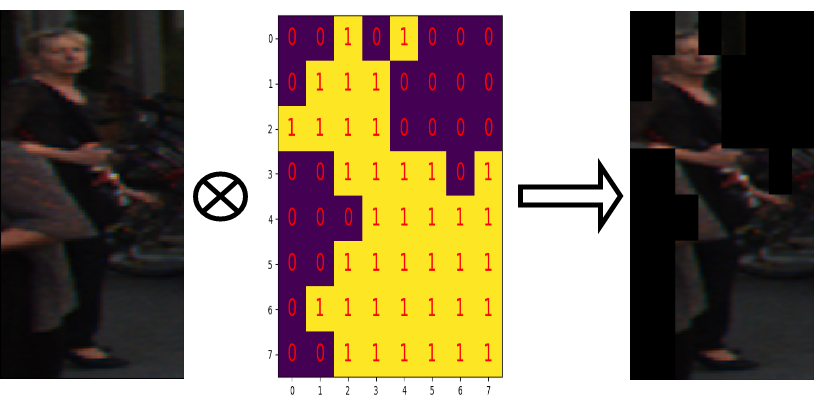}
\caption{}
\label{fig:2}
\end{subfigure}
\caption{(a) Illustration of foreground masking layer procedure: Step 1: calculating the median of nonzero values in each grid; Step 2: obtaining a mask by Equation \ref{m1} ($d_{min}= 6.8, d_{max} = 9.7$ in this example); Step 3: applying the mask to the feature maps. (b) A qualitative example of a foreground mask and its application to the original image. The bottom left background and the top left and right background are masked.}
\label{fig:mask_proc} 
\end{figure}

In order to identify the foreground and background of images, the depth information of each pixel is necessary. But due to the sparsity of the LIDAR point cloud, most of the depth information in the image plane is unknown. Recently, several approaches were proposed to complete the depth map, e.g., \cite{van2019sparse}, \cite{ma2018self}. Unfortunately, they typically have high GPU memory usage, and thus are not suitable for our implementation. Instead we introduce a light-weight method to take advantage of the sparse depth information.

\begin{table*}[ht]
\begin{center}
\begin{tabular}{cc|cccccc}
    & & \multicolumn{3}{c}{$\text{AP}_{\text{3D}}$($\%$)} & \multicolumn{3}{c}{$\text{AP}_{\text{BEV}}$($\%$)}\\
     \hline 
    Method & Class &E & M & H  & E & M & H \\
    \hline
     AOVD-FPN\cite{ku2018joint} &\multirow{4}{*}{Car} & $\mathbf{81.94}$ &$\mathbf{71.88}$ & $\mathbf{66.38}$ &$88.53$&$83.79$&$\mathbf{77.90}$   \\ 
     MV3D \cite{chen2017multi}& & $71.09$ &$62.35$ & $55.12$&$86.02$&$76.90$&$68.49$ \\
     F-PointNets \cite{qi2017pointnet}& & $81.20 $ & 	$70.39$ & 	$62.19 $&$\mathbf{88.70}$ & 	$\mathbf{84.00}$ & 	$75.33$ \\
    Ours & & $72.24$  &	$64.20$ & 	$57.20$ &$85.95$ & 	$77.86$ & 	$76.93$ \\
    \hline
    AOVD-FPN\cite{ku2018joint} &\multirow{4}{*}{Pedestrian} & $50.80$&$ 42.81$ & 	$\mathbf{40.88}$&$\mathbf{58.75}$ &	$\mathbf{51.05}$ & 	$\mathbf{47.54} $   \\ 
     MV3D \cite{chen2017multi}& & - &- &  - & - & - & - \\
     F-PointNets \cite{qi2017pointnet}& & $\mathbf{51.21}$ &	$\mathbf{44.89}$ &	$40.23$&$58.09$ &	$50.22$ &$47.20$ \\
     Ours & & $48.26$ & $40.97$& $35.74$&$52.24$& 	$44.40$& 	$43.24$ \\
    \hline
    AOVD-FPN\cite{ku2018joint} &\multirow{4}{*}{Cyclist} & $64.00$ &$ 	52.18$ &$ 	46.61$&$68.09$ & 	$57.48 $ &	$50.77 $   \\ 
     MV3D \cite{chen2017multi}& & - & -  & - & - &-&- \\
     F-PointNets \cite{qi2017pointnet}& & $\mathbf{71.96}$ & $\mathbf{56.77}$ &$\mathbf{50.39}$& $\mathbf{75.38}$ &	$\mathbf{61.96}$ & 	$\mathbf{54.68} $ \\
    Ours & & $67.66$ & 	$49.89$ & 	$42.23$&$69.68$ &	$58.21$ & 	$50.14$ \\
    \hline
\end{tabular}
\end{center}
\caption{A comparison of the performance of MLOD with current state-of-art 3D object detectors }
\label{tab:test_set}
\end{table*}

Fig. \ref{fig:mask_proc} presents the procedure of the foreground masking layer. First, the layer crops and resizes the (sparse) depth map using front-view 2D bounding boxes, projected from the 3D proposals.  For computational convenience, the resized depth map is $n$ times the $k \times k$ size of the cropped image feature map. Since the depth information is discontinuous in front view, we use nearest neighbour interpolation algorithm to obtain the resized depth map. Then the $nk\times nk$ depth map is split equally into a $k \times k$ grid. Thus each grid cell represents the depth information of the corresponding pixel in the $k\times k$ image feature map. The layer calculates the median $m_{ij}$ of the nonzero depth values in each grid cell, as zero value means no LIDAR point information for this pixel. Note that all depth values in a grid cell may be zero, due to the sparsity of point cloud. Since far objects have fewer projected LIDAR points, some parts of these objects do not have any depth information. Thus, to preserve the image features that are inside the 3D bounding box or have no depth information, we set the foreground mask as
\begin{equation}
    \textit{Mask}_{ij} = \begin{cases} 1 \quad \text{if }  m_{ij} \in [d_{min} -\epsilon_1, d_{max} +\epsilon_1] \cup [0, \epsilon_2]\\
0 \quad \text{otherwise,}
\end{cases}
\label{m1}
\end{equation}
where $d_{max}$ and $d_{min}$ are the maximum and minimum depth value of a 3D bounding box, respectively. $\epsilon_1$ and $\epsilon_2$ are small buffers to absorb the uncertainty of 3D proposals and point cloud.

\subsection{Multi-view Header}
\begin{figure}[ht]
\centering
\includegraphics[width=250pt, height = 180pt]{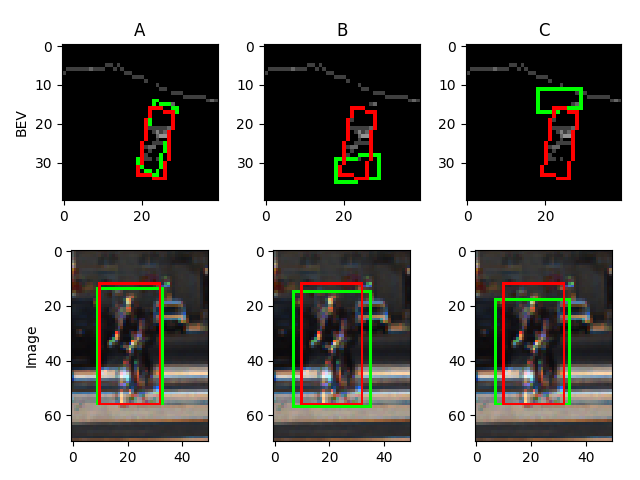}
\caption{Examples of IoU in different views. The pictures show the projection of 3D bounding boxes (proposals A,B,C in green and the ground truth in red) onto ground plane (BEV) and image. The IoU of proposals B and C is less than 0.3 in BEV, but is larger than $0.7$ in image view. Hence proposals B and C are negative in BEV and positive in front view.}
\label{fig:label} 
\end{figure}

In the current multi-view 3D object detection methods, the labels of proposals are assigned based on the IoU in BEV. But the IoU in front view can be significantly different than that in BEV. Fig. \ref{fig:label} shows an example that a 3D bounding box is assigned to a negative label, but has $\text{IoU} > 0.7$ in image view. When object detectors are trained on labels assigned based on IoU only in BEV, the performance of (front-view) image channel is degraded.

We propose a multi-view detection header to avoid the decay of RGB image features. Fig. \ref{fig:multi-view_header} shows the header network structure. The key idea is to add an extra output layer to each channel before the (Concat) fusion layer. Each of the two outputs feeds into a corresponding sub-output loss. Each sub-output loss is calculated using the labels assigned by IoU in the corresponding channel's view, i.e.
\[
\begin{split}
    L_{sub-cls} &= \frac1{N} \sum_i L_{cls}(y^{img}_i, \hat{y}^{img}_i) \\
    &+ \frac1{N} \sum_i L_{cls}(y^{bev}_i, \hat{y}^{bev}_i)\\
\end{split}
\]
\[
\begin{split}
L_{sub-reg} &= \frac1 {N^{img}_{p}} \sum_i I[\hat{y}^{img}_i > 0]L_{reg}(s^{img}_i, \hat{s}^{img}_i) \\
    &+ \frac1{N^{bev}_{p}} \sum_i I[\hat{y}^{bev}_i > 0] L_{reg}(s^{bev}_i, \hat{s}^{bev}_i).\\
\end{split}
\]
$I[\cdot > 0]$ is the indicator function to select the positive proposals. $N$, $N^{img}_{p}$ and $N^{bev}_{p}$ are the number of total samples, positive samples in image view and BEV, respectively. $y^{img}_i$ and $y^{bev}_i$ are the classification scores for proposal $i$ obtaied from image and BEV branch, respectively,  and $\hat{y}^{img}_i$ and $\hat{y}^{bev}_i$ are the corresponding ground-truth labels. The predicted corner offsets for each branch are $s^{img}_i$ and $s^{bev}_i$, and the corresponding ground truth lables are $ \hat{s}^{img}_i$ and $\hat{s}^{bev}_i$.

We use a multi-task loss to train our network. The loss function of the detection network is defined by Eq. \ref{loss},
\begin{equation}
\begin{split}
L &= \frac{\lambda_{cls}}{N} \sum_i L_{cls}(y^{fusion}_i, \hat{y}^{bev}_i)    \\
  &+ \frac{\lambda_{reg}}{N^{bev}_{p}} \sum_i I[\hat{y}^{bev}_i > 0] L_{reg}(s^{fusion}_i, \hat{s}^{bev}_i) \\
  &+ \frac{\lambda_{ang}}{N^{bev}_{p}} \sum_i I[\hat{y}^{bev}_i > 0]L_{ang}(a^{fusion}_i, \hat{a}^{bev}_i) \\
  &+ \lambda_{sub-cls}L_{sub-cls} + \lambda_{sub-reg}L_{sub-reg}.
\end{split}
    \label{loss}
\end{equation}

We use smooth L1 loss for 3D bounding box offset and orientation rotation regression, and cross-entropy loss for classification. 
$\lambda$ are the hyperparameters to balance the different loss terms. The sub-output losses can be considered as a kind of regularization on the network.

\section{Training}
\label{sec4}
KITTI benchmark \cite{geiger2012we} uses different IoU thresholds for the car class ($>$0.7) and the pedestrian and cyclist classes ($>$0.5). Hence, following \cite{ku2018joint}, we train two networks, one for cars, another for pedestrians and cyclists. The RPN network and the detection header are trained jointly using mini-batches with 1024 ROIs. We use ADAM \cite{kingma2014adam} optimizer with an exponentially decayed learning rate initialized to $0.0001$. For the car network, we apply a decay factor of $0.1$ every $100$K iterations. For the pedestrian and cyclist network, we apply a decay factor of $0.5$ every $20$K iterations. Image feature extractor loads pre-trained ImageNet \cite{deng2009imagenet} weights.  The weights of the BEV feature extractor are initialized by Xavier uniform initializer \cite{glorot2010understanding}.

\subsection{Mini-batch Settings}
A car proposal is marked as positive in top-down/front view if the BEV/image IoU with a ground truth object is larger than $0.65$/$0.7$, respectively. It is marked negative if its BEV/image IoU is less than $0.55$/$0.5$, respectively. A positive pedestrian or cyclist proposal has at least $0.45$/$0.6$ IoU in BEV/image view, respectively. A negative sample has no more than $0.4$/$0.4$ IoU in BEV/image view, respectively. For mini-batches, we first select $1024$ samples consisting of both positive ROIs and negative ROIs with highest RPN scores in top-down view, then pick ROIs which are positive or negative in front view. 

\subsection{3D Box Encoding}
There are many ways to encode 3D boxes (e.g., \cite{ku2018joint}, \cite{chen2017multi}, \cite{song2016deep}). To reduce the number of parameters and keep physical restrictions, we follow the encoding method from \cite{ku2018joint}, where the 3D bounding box is represented as four corners on X-Y plane and  the  top  and  bottom  corner height offsets  from  the  ground plane. 

\begin{figure*}[ht]
\centering
\includegraphics[width=460pt,height=200pt]{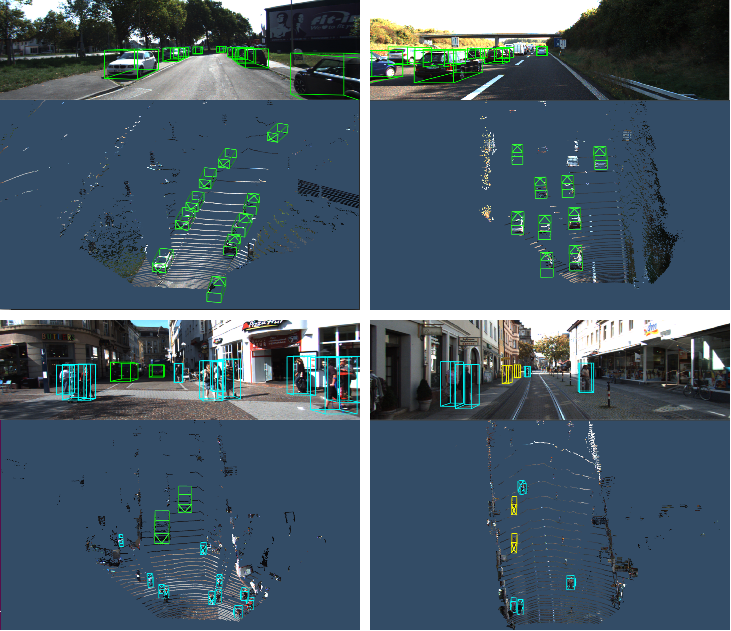}
\caption{Qualitative results of MLOD. In each image, detected cars are in green, pedestrians are in blue, and cyclists are in yellow.}
\label{fig:qua} 
\end{figure*}

\subsection{Data Augmentation}
Data augmentation is an important technique for increasing the number of training instances and reducing overfitting. Two augmentation methods, flipping and PCA jittering \cite{krizhevsky2012imagenet}, are implemented in our network training. The point clouds and images are flipped along the x-axis. PCA jittering alters the intensities of the RGB channels in training images. PCA decomposition is applied to the set of RGB pixel values of the whole set of training images. Then Gaussian random noise is added to the principle components of images.

\section{Experiments}
\label{sec5}

\subsection{KITTI Dataset and Metrics}
We evaluate MLOD on the 3D detection tasks for cars, pedestrians, and cyclists of the KITTI Object  Detection  Benchmark \cite{geiger2012we}. The 3D object detection dataset of KITTI contains $7,481$ training frames and $7,518$ testing frames. The frames contain target-class objects categorized into into three difficulty levels: easy (E), moderate (M), and hard (H), based on the occlusion level, maximum truncation and minimum bounding box height. Since no official validation set is provided, the labelled $7,481$ frames are split into a training set and a validation set at $1:1$ ratio, similar to \cite{ku2018joint} and \cite{chen2017multi}. 

\subsection{Accuracy}
To evaluate the performance of MLOD, we present the Average Precision (AP) results over the validation set and the KITTI test set in Table \ref{tab:val_result} and \ref{tab:test_set}, respectively. MLOD outperforms two other state-of-the-art multi-view object detectors on the validation set. However, our method perfroms worse than AVOD on the KITTI test set. This may caused by the different ground planes used in MLOD and AVOD. The evaluation shows that our method can reach the current state of art result, however. 

\begin{table}[ht]
\begin{center}
\begin{tabular}{ccccc}
Method & Cars & Pedestrians & Cyclist \\
\hline
MV3D \cite{chen2017multi} &$72.4$ & - & - \\
AVOD \cite{ku2018joint}&$\mathbf{74.4}$ & $58.8$ & $49.7$\\
Ours & $74.1$ & $\mathbf{63.9}$ & $\mathbf{54.6}$
\end{tabular}
\end{center}
\caption{A comparison of $\text{AP}_{\text{3D}}$ from MLOD and current state-of-art 3D object detectors on validation set at the moderate difficulty. }
\label{tab:val_result}
\end{table}

\subsection{Effects of Multi-view Header}
\begin{figure}[ht]
\centering
\includegraphics[width=250pt]{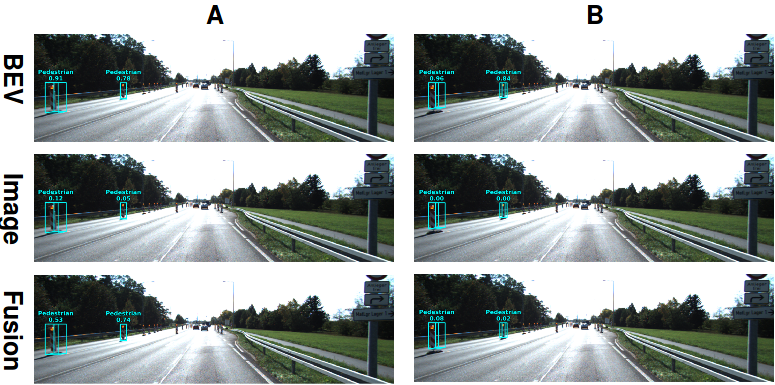}
\caption{Examples of the effects of various $\lambda$ settings. Column A: $\lambda_{sub-cls}/\lambda_{cls} = 0.001$; Column B: $\lambda_{sub-cls}/\lambda_{cls} = 1$} 
\label{fig:header_effect_pics} 
\end{figure}

To evaluate the effects of the multi-view header, we compare the AP($\%$) of MLOD with different $\lambda_{sub-cls}$ settings in Table \ref{tab:header_effect} on the validation set. When $\lambda_{sub-cls}/\lambda_{cls} = 0.001$, the fusion channel, with BEV labelled samples, dominates the network training, such that the sub-channel losses are ignorable. The multi-view header is shown to provide significant performance gains for image channel, ranging from $5\%$ to $20\%$, however. The final detection AP achieves an increase of $6.7\%$, $5.2\%$ and $4.5\%$ in AP for Pedestrians Easy, Moderate, and Hard classes, respectively. Figure \ref{fig:header_effect_pics} shows an example of the effects of multi-view header. Note when $\lambda_{sub-cls}/\lambda_{cls} = 1$, the image channel correctly assigns score of $0.0$ to the pedestrian false positives from LIDAR BEV. 

\begin{table}[ht]
\begin{center}
\begin{tabular}{cccccccc}
    & & \multicolumn{3}{c}{Pedestrians} & \multicolumn{3}{c}{Cyclist}\\
    \cline{3-8} \\
    $\frac{\lambda_{sub-cls}}{\lambda_{cls}}$ & Branch & E & M & H & E & M & H \\
    \hline
    \multirow{2}{*}{$0.001$} &Fusion & $65.2$ &$58.7$ & $51.7$&$71.5$&$53.6$&$47.5$   \\ \cline{2-8} 
     &Image & $53.3$ & $47.3$& $41.5$ & $39.5$& $23.6$& $22.8$ \\
    \hline
    \multirow{2}{*}{$1$} &Fusion & $71.9$ &$63.9$&$56.2$&$73.5$&$54.6$&$52.7$   \\ \cline{2-8} 
     &Image & $59.4$ &$52.8$&$49.7$&$59.2$&$40.2$&$38.5$ \\
\end{tabular}
\end{center}
\caption{$\text{AP}_{\text{3D}}$ from MLOD with different $\lambda$ settings, evaluated on the validation set. Since the image channel lacks depth information, it is difficult to predict the 3D bounding box from it. To facilitate the comparison, results from fusion and image channel use the same 3D bounding boxes. Thus, the shown results reflect only the variation of classification results.}
\label{tab:header_effect}
\end{table}

\subsection{Effects of Foreground Mask Layer}
Table \ref{tab:mask_effect} shows how the mask component affects the performance of MLOD.

\begin{table}[ht]
\begin{center}
\begin{tabular}{ccccccc}
     & \multicolumn{3}{c}{Pedestrians} & \multicolumn{3}{c}{Cyclist}\\
    \hline \\
       & E & M & H & E & M & H \\
    \hline
    With masks  & $\mathbf{71.9}$ &$\mathbf{63.9}$&$\mathbf{56.2}$&$73.5$&$\mathbf{54.6}$&$\mathbf{52.7}$   \\ 
    \hline
    W/o masks & $69.1$ &$61.4$&$53.6$&$\mathbf{74.1}$&$54.2$&$52.5$ \\
\end{tabular}
\end{center}
\caption{Effects of a foreground mask layer.}
\label{tab:mask_effect}
\end{table}

\subsection{Qualitative Results}
Some qualitative results in 3D and image space are presented in Fig. \ref{fig:qua}.

\section{Conclusion}
\label{sec6}
We have proposed a multi-view based 3D object detection model for autonomous driving scenarios. In order to obtain the image features inside 3D bounding box proposals, a foreground mask layer is introduced. Furthermore, training with the multi-view labelled data prevents the decay of the image channel, such that the proposed detector can provide better  classification and localization results. Evaluated on KITTI detection dataset, our method achieves state of  the  art  benchmark  results.

\bibliographystyle{IEEEtran}
\bibliography{main}
\end{document}